\documentclass[conference]{IEEEtran}
\IEEEoverridecommandlockouts
\usepackage{cite}
\usepackage{amsmath,amssymb,amsfonts}
\usepackage{algorithmic}
\usepackage{textcomp}
\usepackage{xcolor}
\usepackage{balance}
\usepackage{multirow}
\usepackage{adjustbox}
\usepackage{svg}
\usepackage{lipsum}
\usepackage{multicol}
\usepackage{graphicx}
\usepackage{rotating}
\usepackage{balance}
\usepackage{comment}
\usepackage{hyperref}
\hypersetup{
    colorlinks=true,    
    linkcolor=black,    
    citecolor=black,     
    urlcolor=black       
}
\def\BibTeX{{\rm B\kern-.05em{\sc i\kern-.025em b}\kern-.08em
    T\kern-.1667em\lower.7ex\hbox{E}\kern-.125emX}}
\begin{document}
\title{MIS-AVoiDD: \textbf{M}odality \textbf{I}nvariant and \textbf{S}pecific Representation for \textbf{A}udio-\textbf{Vi}sual \textbf{D}eepfake \textbf{D}etection}
\author{\IEEEauthorblockN{Vinaya Sree Katamneni and Ajita Rattani}
\IEEEauthorblockA{\textit{Dept. of Computer Science and Engineering}\\
\textit{University of North Texas, Denton, USA}\\
        \text{vinayasreekatamneni@my.unt.edu};
        \text{ajita.rattani@unt.edu}}}        
\maketitle

\begin{abstract}
Deepfakes are synthetic media generated using deep generative algorithms and have posed a severe societal and political threat. 
Apart from facial manipulation and synthetic voice, recently, a novel kind of deepfakes has emerged with either audio or visual modalities manipulated. 
In this regard, a new generation of multimodal audio-visual deepfake detectors is being investigated to collectively focus on audio and visual data for multimodal manipulation detection. Existing multimodal (audio-visual) deepfake detectors are often based on the fusion of the audio and visual streams from the video.
Existing studies suggest that these multimodal detectors often obtain equivalent performances with unimodal audio and visual deepfake detectors. 
We conjecture that the heterogeneous nature of the audio and visual signals 
creates distributional modality gaps and poses a significant challenge to effective fusion and efficient performance. 
In this paper, we tackle the problem at the representation level to aid the fusion of audio and visual streams for multimodal deepfake detection. Specifically, we propose the joint use of modality (audio and visual) invariant and specific representations. This ensures that the common patterns and patterns specific to each modality representing pristine or fake content are preserved and fused for multimodal deepfake manipulation detection.
Our experimental results on FakeAVCeleb and KoDF audio-visual deepfake datasets suggest the enhanced accuracy of our proposed method over SOTA unimodal and multimodal audio-visual deepfake detectors by $17.8$\% and $18.4$\%, respectively. Thus, obtaining state-of-the-art performance.



\end{abstract}


%
\section{Introduction}
\label{sec:intro}

With the advances in deep generative models~\cite{10.1145/3625547}, synthetic audio and visual media have become so realistic that they are often indiscernible from authentic content for human eyes. However, synthetic media generation techniques used by malicious users to deceive pose a severe societal and political threat~\cite{Hwang2020, citron}. 
In this context, visual (facial) deepfakes are generated using facial forgery techniques that depict human subjects with altered identities, malicious actions, and facial attribute manipulation. However, the recent advancements in deepfake generation techniques have also resulted in cloned human voices in real time. Human voice cloning~\cite{chintha2020recurrent,pianese2022deepfake} is a neural-network-based speech synthesis method that takes an audio sample of the target person and text as input and generates a high-quality speech of the target speaker's voice. These audio and visual deepfakes have been employed to attack authentication systems, impersonate celebrities and politicians, and defraud finance. As a countermeasure, several unimodal audio and visual deepfake detectors have been proposed~\cite{9157215,9578910,9360904, journals/corr/abs-1003-4083,chintha2020recurrent,hamza2022deepfake,pianese2022deepfake}.
With the recent advances in generation techniques and the ability to easily forge videos with lip-synced synthetic audios, a novel kind of \textbf{multimodal} deepfakes has emerged with either one or both audio and visual modalities manipulated~\cite{10.1145/3476099.3484315, ILYAS2023110124, 10.1145/3394171.3413570, 10.1145/3394171.3413700, 9350195, 10095247, 9980296, Cozzolino2022AudioVisualPD,  cheng2022voice, feng2023self,  raza2023multimodaltrace, chen2023npvforensics}. The existing unimodal deepfake detectors are primarily designed to detect a single type of manipulation, either visual or acoustic. Consequently, a new generation of multimodal\footnote{The terms multimodal and audio-visual deepfake detectors are used interchangeably in this paper.} deepfake detectors are being investigated to detect audio and visual manipulations collectively. Existing audio-visual deepfake detectors are often based on the fusion of audio and visual streams using multimodal convolutional neural networks (CNNs)~\cite{zhou2021joint, 9350195, 10.1145/3394171.3413570}, and ensemble-based voting schemes~\cite{10.1145/3476099.3484315}. 



\begin{figure}
\centering
    \includegraphics[width=0.5\textwidth]{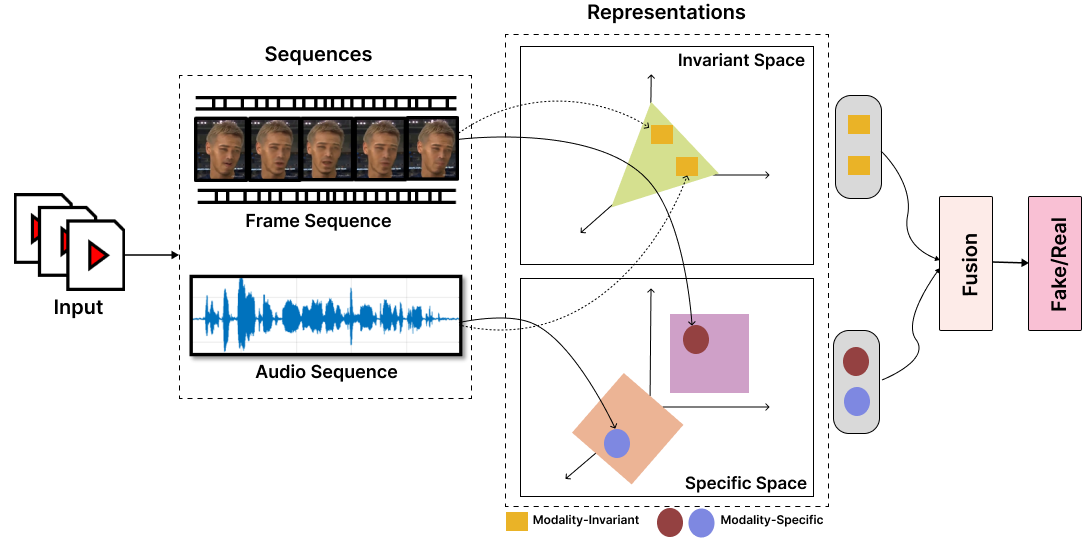}
    \caption{The schema of our proposed 
    audio-visual deepfake detector based on multimodal representation learning through modality invariant and -specific subspaces.  
    The learned feature representations are fused for multimodal deepfake detection.} 
    \label{fig:img1}
\end{figure}

Current studies~\cite{10.1145/3476099.3484315, zhou2021joint, 10.1145/3394171.3413570,katamneni_nadimpalli_rattani_2023, 10.1145/3394171.3413700} suggest that existing audio-visual deepfake detectors often obtain similar performance or marginal improvement over unimodal audio and visual deepfake detectors. This is due to the heterogeneous nature of the audio and visual signals that create a modality gap and pose a significant challenge in information fusion and accurate multimodal deepfake detection. 

The \textbf{aim} of this paper is to propose a novel method of audio-visual deepfake detection based on the fusion of modality invariant and modality-specific feature representation for audio and visual streams. This is facilitated through a novel framework that utilizes two distinct subspaces (modality invariant and specific subspaces) to project each modality. The first subspace captures commonalities across the two modalities and the other subspace captures the unique characteristics of each modality.  This ensures that the common patterns and patterns specific to each modality representing pristine or fake content are preserved and fused for efficient multimodal deepfake manipulation detection. 

Figure~\ref{fig:img1} illustrates an overview of the proposed approach which involves learning audio and visual representations through modality-invariant and modality-specific subspaces. These modality invariant and specific representations are then fused for deepfake detection.

Accordingly, the \textbf{contributions} of this paper are as follows:
\begin{enumerate}

\item A novel multimodal deepfake detector, named MIS-AVoiDD, based on the fusion of modality invariant and specific feature representations for audio and visual streams.

\item Comparison of our proposed model with the published work on audio, visual, and audio-visual (multimodal) deepfake detectors.



\item Evaluation on publicly available audio-visual FakeAVCeleb~\cite{DBLP:journals/corr/abs-2108-05080} and KoDF~\cite{Kwon_2021_ICCV} deepfake datasets.


\end{enumerate}


This paper is summarized as follows: Section~\ref{relatedwork} discusses the related work on audio-visual deepfake detection and multimodal fusion. The proposed approach is discussed in Section~\ref{approach}. Section~\ref{experiment} discusses the datasets, the evaluation metrics, and the results. The ablation study conducted is detailed in the Section~\ref{ablation}. Conclusion and future research directions are discussed in Section~\ref{conclusion}.

\section{Related Work}
\label{relatedwork}

\noindent \textbf{Audio-Visual Deepfake Detection:} 
Study in~\cite{10.1145/3476099.3484315} assembled FakeAVCeleb dataset consisting of audio and visual deepfakes and benchmarked various audio-visual deepfake detectors based on ensemble-based voting scheme and multimodal CNN based on feature concatenation. 
In~\cite{zhou2021joint}, the authors proposed a two-plus-one-stream model that separately modeled the audio and visual streams, including a sync stream to model the synchronization patterns between the two modalities by concatenating audio and visual streams using an attention mechanism.

In~\cite{10.1145/3394171.3413570}, a novel approach is proposed that simultaneously exploits the audio and visual (face) modalities and the perceived emotion extracted from both modalities to detect deepfakes. To facilitate this, the Siamese network was trained on modality and perceived emotion embedding vectors extracted from the face and speech of the subject using triplet loss. 

In~\cite{10.1145/3394171.3413700}, the dissimilarity between feature embeddings from the audio and visual modalities, obtained using the ResNet model, was utilized for deepfake detection. The method combined cross-entropy loss for classification with contrastive loss to model inter-modality similarity. 

The study in~\cite{9980296} proposed an audio-visual lipreading-based model fine-tuned on the embeddings of lip sequences from visual modality and synthetic lip sequences, generated from the audio using the Wav2lip model, for deepfake detection. 
The absolute difference between the real and synthetic lip sequence embeddings was used for deepfake detection.
Recently, in~\cite{yang2023avoid} an audio-visual model, AVoiD-DF model, is proposed that consists of a temporal-spatial encoder to embed temporal-spatial information and a multimodal joint-decoder to fuse multimodal features and jointly learn inherent relationships. This is followed by a cross-modal classifier to detect manipulations with inter-modal and intra-modal disharmony.

Very recently, in~\cite{raza2023multimodaltrace}  novel audio-visual patch mixer, Multimodaltrace, is proposed which fuses learned channels from audio and visual modalities, independently, using intra-modality mixer layer and jointly using inter-modality mixer layer. These features are learned through a shared MLP block to learn patterns between patches followed by the patterns within channels.\\ 

\noindent \textbf{Multimodal Fusion:} 
Several methods have been proposed for the fusion of multiple sources of information, such as multi-sensor data, multi-classifiers, and multi-modalities using feature concatenation and ensemble learning-based hard and soft voting schemes~\cite{10.1007/978-1-4471-0833-7_2}, typically used for the performance enhancement of a pattern recognition classifier. 
Apart from the aforementioned, common subspace representations have been used for multimodal feature representation using translation-based models~\cite{sulubacak2020multimodal} to translate one modality to another and canonical correlation analysis~\cite{sahbi2018learning} to learn cross-modal correlations.
Within the regime of subspace learning, factorized representation such as learning generative-discriminative factors~\cite{tsai2018learning} of multimodal data have been proposed for the heterogeneous modalities. The shared-private representation learning via multi-view component analysis including latent variable models (LVMs)~\cite{8840977} with separate shared and private latent
variables have also been used for multimodal fusion. 

\begin{figure*}
\centering
    \includegraphics[width=1.01\textwidth]{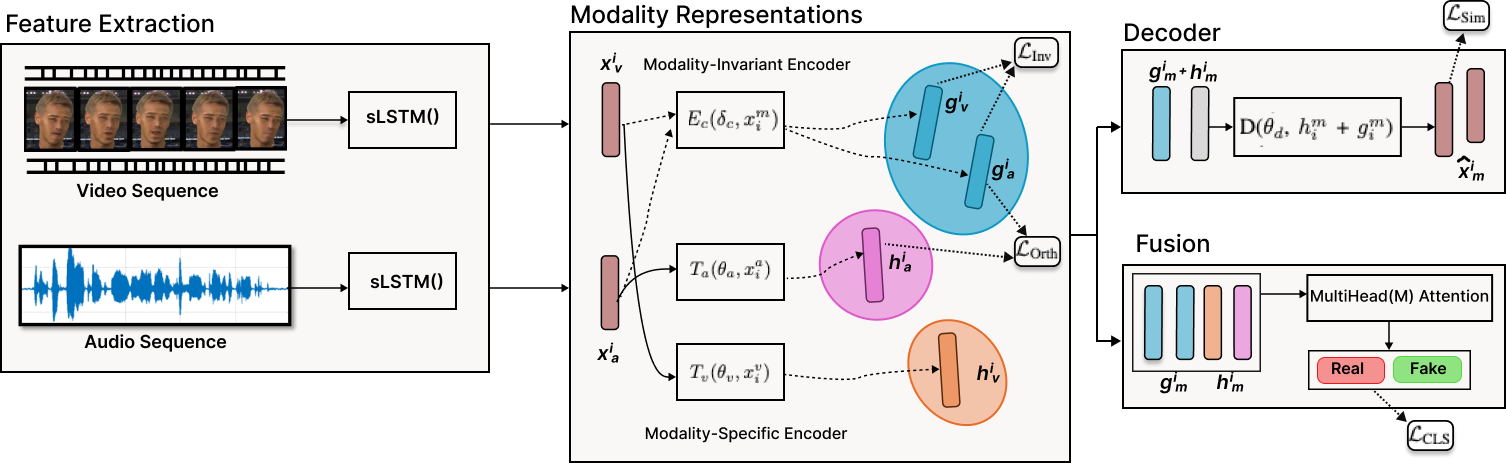}
    \caption{MIS-AVoiDD: Modality Invariant and Specific Representation for Audio-Visual Deepfake Detection projects sequence-level feature representations of each modality (audio and visual) into two distinct subspaces; modality-invariant and modality-specific features. The hidden sequence-level representations are fused for each sequence in a video. The final deepfake detection is done by averaging the classification loss for each of the $N$ sequences in a video.}
    \label{fig:img2}
\end{figure*}

\section{MIS-AVoiDD: Approach}
\label{approach}
\subsection{\textbf{{Problem Formulation}}}
We aim to develop an efficient audio-visual deepfake detection system that can leverage multimodal signals namely, audio and visual. Our approach focuses on developing effective modality representations that can be used to improve the process of multimodal fusion. To achieve this, we propose a novel framework called MIS-AVoiDD explained as follows. 

For a given video $D$ that contains $N$ sequences, where each sequence $d_i$ is composed of $2$ different modalities, the formulation can be represented as follows:
\begin{equation}
\label{eq1}
D = \{d_i\}_{i=1}^N  \quad ; \quad d_i = \{x_{i}^{a}, x_{i}^{v}\}
\end{equation}

Here, each sequence $d_i$ consists of two modalities represented by $x_{i}^{m}$ for $m \in \{a,v\}$ for audio and visual modality.  Typically, the learned representations for these modalities exist in different feature spaces, making it challenging to compare their similarities directly. Our proposed model aims to learn two distinct types of features for each modality: modality-invariant features which remain invariant across modalities, and modality-specific features which capture the unique characteristics of each modality. By jointly learning these modality-invariant and modality-specific features using our framework, we can obtain more comprehensive representations for each data instance.  Figure~\ref{fig:img2} illustrates the step-by-process involved in our proposed MIS-AVoiDD model. The detailed formulation of our method can be found in the subsequent sections.


\subsection{\textbf{Modality Representation}}
\label{representation}
For each input sequence \(d_i\), as seen in Figure~\ref{fig:img2}, we have two modalities represented by audio and visual sequences \(x_{i}^{a}\) and \(x_{i}^{v}\) respectively. To capture the inherent characteristics of each modality, we utilize specific feature encoders \(T_a\) and \(T_v\) tailored for each modality. These encoders transform the \(x_{i}^{a}\) and \(x_{i}^{v}\) into hidden vectors \(h_{i}^{a}\) and \(h_{i}^{v}\) using learnable parameters \(\theta_a\) and \(\theta_v\). The hidden vectors \(h_{i}^{a}\) and \(h_{i}^{v}\) represent modality-specific features.
\begin{equation}
\label{eq2}
     h_{i}^{a} = T_a(\theta_a, x_{i}^{a}) \quad ; \quad  h_{i}^{v} = T_v(\theta_v, x_{i}^{v}) 
\end{equation} 

To capture the invariant characteristics of each modality, we utilize invariant feature encoder \(E_c\) with shared learnable parameter $\delta_c$ as seen in Figure~\ref{fig:img2}. The encoder is a feed-forward neural layer that transforms the \(x_{i}^{a}\) and \(x_{i}^{v}\) into hidden vectors \(g_{i}^{a}\) and \(g_{i}^{v}\). The hidden vectors \(g_{i}^{a}\) and \(g_{i}^{v}\) represent modality-invariant features.

\begin{equation}
\label{eq3}
    g_{i}^{a} = E_c(\delta_c, x_{i}^{a}) \quad ; \quad g_{i}^{v} = E_c(\delta_c, x_{i}^{v}) 
\end{equation} 

Where, \(E_c\) maps the feature representation \(x_{i}^{m}\) to the universal feature space, while \(T_m\) maps \(x_{i}^{m}\) into the modality-specific feature space where $m \in \{a, v\}$ for audio and visual modalities. During the learning process, \(h_{i}^{m}\) and \(g_{i}^{m}\) are obtained by training the entire network with a combination of different constraints proposed in our approach. To generate the specific and invariant representations, encoders use a simple feed-forward neural layer; $E_c$ shares the parameters $\delta_c$ across audio and visual modalities, while $T_a$ and $T_v$ have separate parameters $\theta_a$ and $\theta_v$ for each modality.

\subsection{\textbf{Modality Fusion}}
Once the audio and visual modalities are projected into their respective representations \{$h_{i}^{a}, h_{i}^{v}, g_{i}^{a}, g_{i}^{v}$\} (refer Eq.~\ref{eq2} and $3$), they are fused into a joint vector to make downstream deepfake detection (predictions) as depicted in Figure~\ref{fig:img2}. To accomplish this, a fusion mechanism is designed that first utilizes self-attention, based on the transformer, followed by a concatenation of all four transformed modality vectors \{$\overline{h}_{i}^{a}, \overline{h}_{i}^{v}, \overline{g}_{i}^{a}, \overline{g}_{i}^{v}$\}. The transformer uses an attention module defined as a scaled dot-product function, where query, key, and value matrices are denoted as $Q$, $K$, and $V$, respectively. The transformer computes multiple parallel attentions each of which is called a head. 
\begin{equation}
    \label{eq4}
    A(Q,K,V) = \text{softmax}\left(\frac{{Q_hK_h^\top}}{{\sqrt{{d_k}}}}\right)
\end{equation}  

In this equation, \(A\) represents the attention scores for the \(h\)-th attention head, \(Q_h\) represents the projected queries, \(K_h\) represents the projected keys, and \(d_k\) represents the dimensionality of the keys and queries. The softmax function is applied to normalize the attention scores.
\begin{equation}
    \label{eq5}
    \text{head}_i = \text{Attention}(Q \cdot W_{i}^q, K \cdot W_{i}^k, V \cdot W_{i}^v)
\end{equation}

In this equation, \(\text{head}_i\) represents the output of the \(i\)-th attention head, \(Q\), \(K\), and \(V\) are matrices representing the query, key, and value vectors, respectively, and \(W_{i}^q\), \(W_{i}^k\), and \(W_{i}^v\) are weight matrices for the query, key, and value projections, respectively. The \(\text{Attention}\) function represents the softmax-scaled dot product attention mechanism.

The four representations are stacked into a matrix $M = [h_{i}^{a}, h_{i}^{v}, g_{i}^{a}, g_{i}^{v} ]$ and multi-headed self-attention is applied to make each vector aware of the fellow cross-modal and cross-subspace representations. The transformer generated matrix $\overline{M} = [\overline{h}_{i}^{a}, \overline{h}_{i}^{v}, \overline{g}_{i}^{a}, \overline{g}_{i}^{v}]$ as:
\begin{equation}
    \label{eq6}
    \overline{M} = \text{MultiHead}(M; \Theta^{att}) \\
                = (\text{head}_1 \oplus \ldots \oplus \text{head}_n)W^o
\end{equation} 

Where \(\overline{M}\) represents the transformed matrix, \(\text{MultiHead}\) is the function that applies multi-head attention, \(M\) is the input matrix stack of four representations, \(\Theta^{att}\)=\(\{W^q, W^k, W^v, W^o\}\), \(\text{head}_i\) represents the output of the \(i\)-th attention head, \( n \) represents the number of attention heads and \(\oplus\) denotes the concatenation operation. Finally, \(W^o\) is the weight matrix applied to the concatenated outputs of the attention heads.


{\textbf{Prediction}}
Finally, we construct a joint vector using concatenation and generate task predictions. 
\begin{equation}
\label{eq7}
    h_{out} = [ \overline{h}_{i}^{a} \oplus \overline{h}_{i}^{v} \oplus \overline{g}_{i}^{a} \oplus \overline{g}_{i}^{v} ] \quad;  \quad \hat{y} = G(h_{out}; \Theta_{out})
\end{equation}   
In the above equation, \( h_{out} \) is the concatenated output of the attention heads, 
\( \hat{y} \) represents the predicted task output which is deepfake detection, \( G \) is the function that generates the predictions based on the output, and \( \Theta_{out} \) represents the parameters associated with the prediction function. Note that the prediction is averaged over all the $N$ sequences in a video for deepfake detection.

\subsection{\textbf{Learning}}
\label{regularization}
The model's overall learning process involves minimizing a combined loss function represented as follows:
\begin{equation}
    \label{eq8}
    \mathcal{L} = \alpha \mathcal{L}_{\text{Inv}} + \beta \mathcal{L}_{\text{Orth}} + \gamma \mathcal{L}_{\text{Sim}} + \mathcal{L}_{\text{CLS}} 
\end{equation}

In the equation above, the weights $\alpha$, $\beta$, and $\gamma$ determine the contribution of each regularization component to the overall loss $\mathcal{L}$. Each of these component losses plays a crucial role in obtaining the desired properties of the subspace. All the loss functions are depicted in Figure~\ref{fig:img2}.

\subsubsection{\textbf{$\mathcal{L}_{\text{Inv}}$-Modality-Invariant Loss}}
The invariant loss aims to minimize the discrepancy between invariant representations of each modality. This alignment of common cross-modal features in the universal subspace facilitates improved coherence. To measure this discrepancy, we utilize the Central Moment Discrepancy (CMD) metric, which quantifies the difference between the distributions of two representations by comparing their order-wise moment differences. As the two distributions become more similar, the CMD distance decreases. 
For instance for the modalities $a$ and $v$ CMD is calculated over $g_{i}^{a}$ and $g_{i}^{v}$ from the encoder $E_c(\delta_c)$ (refer Eq.~\ref{eq3}). The invariant loss of all $K^{\text{th}}$ order sample central moments is calculated as follows:
\begin{equation}
\label{eq9}
    \mathcal{L}_{\text{Inv}} = \text{CMD}_{K}(g_{i}^{a}, g_{i}^{v}) 
\end{equation}

\subsubsection{\textbf{$\mathcal{L}_{\text{Orth}}$-Orthogonal Loss}}
To ensure that our model effectively learns distinct aspects of the data for each modality, we introduced additional constraints to regulate the relationship between the modality-specific features $h_{i}^{m}$ (refer Eq.~\ref{eq2}) and the modality-invariant features $g_{i}^{m}$ (refer Eq.~\ref{eq3}) where $m \in \{a, v\}$ for audio and visual modalities. These soft orthogonality constraints aim to prevent the model from learning redundant features between the two modality representations. 
In each sequence, the modality-specific and -invariant features are normalized to have zero mean and unit L$2$ norm. We construct matrices $H_{S}^{m}$ and $H_{I}^{m}$, where each row represents the vectors $h_{i}^{m}$ and $g_{i}^{m}$ (refer Eq.~\ref{eq2} and ~\ref{eq3}) for the respective modality $m \in \{a, v\}$ for audio and visual modalities. The orthogonality constraint between the specific and invariant feature vectors for modality $m$ is computed as the squared Frobenius norm of $H_{S}^{m}$ multiplied by $H_{I}^{m}$ transposed.

Furthermore, we also incorporate orthogonality constraints between the modality-specific vectors themselves. This ensures that the features within each modality remain distinct. The overall difference loss is then calculated by summing the squared Frobenius norms of the pairwise products between $H_{I}^{m}$ and $H_{S}^{m}$, and between $H_{S}^{m}$ and $H_{S}^{m}$ for the modality pairs.
\begin{equation*}
\label{eq10}
 { \mathcal{L}_{\text{Orth}} = \sum_{m \in \{v,a\}}  \|H_{S}^{m} {H_{I}^{m^{T}}\|^2_F} + \sum_{\substack{(m_1,m_2) \in \\ \{(a,v),(v,a)\}}}\|H_{S}^{m_1} H_{S}^{{m_2^{T}}}\|^2_F} \tag{10}
\end{equation*}

\subsubsection{\textbf{$\mathcal{L}_{\text{Sim}}$-Modality-Specific Loss}}
The modality-specific features are designed to capture the unique characteristics that are specific to each modality. To address the risk of learning trivial representations by the modality-specific encoders, we introduce a specific loss that ensures the hidden representations capture the essential details of their respective modalities. We achieve this by reconstructing the modality vector with decoders 
$\hat{x_i}^a$ =D($\theta_d$, $h_{i}^{a}$ + $g_{i}^{a}$) and $\hat{x_i}^v$ =D($\theta_d$, $h_{i}^{v}$ + $g_{i}^{v}$) (refer Eq.~\ref{eq2} and ~\ref{eq3}). 
It is calculated as the mean squared error of the decoded and the encoded representation of each modality using the below equation.

\begin{equation*}
\label{eq11}
    \mathcal{L}_{\text{Sim}} = \frac{1}{2} \sum_{m \in \{v, a\}} \|{x_i}^m - \hat{x_i}^m\|^2_{2} \tag{11}
\end{equation*}


\subsubsection{\textbf{$\mathcal{L}_{\text{CLS}}$-Classification Loss}}
The task-specific loss is employed to evaluate the accuracy of predictions during training. 
We utilized the conventional cross-entropy loss which is calculated as follows:


\begin{equation}
\label{eq12}
    \mathcal{L}_{\text{CLS}} =  -\left[ y \cdot \log(\hat{y}) + (1 - y) \cdot \log(1 - \hat{y}) \right] \tag{12}
\end{equation} 

Here, $y$ represents the true label or target value, and $\hat{y}$ represents the predicted value or probability. The average of this loss over the batch of $N$ sequences in a video is the task-specific loss and is used for deepfake detection of the video.

\section{Experimental Validations and Implementation details}
\label{experiment}
\subsection{{{\textbf{Dataset}}}}
\label{metrics}
\begin{figure}
    \centering
    \includegraphics[width=0.49\textwidth]{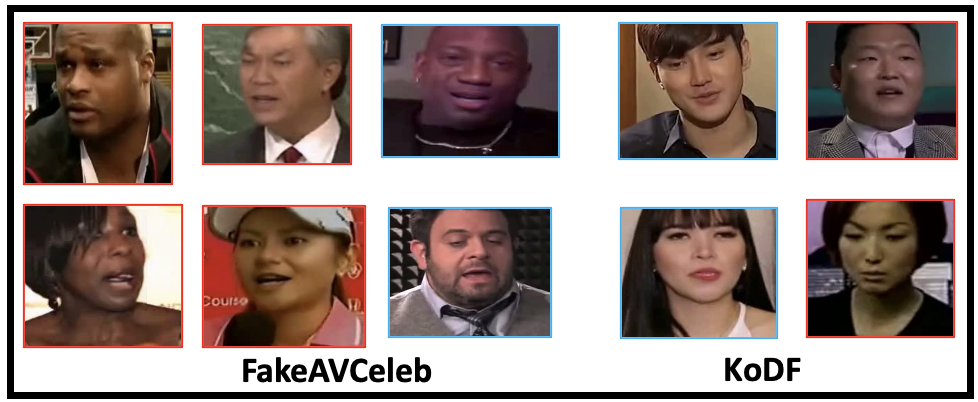}
    \caption{Sample frames from FakeAVCeleb~\cite{DBLP:journals/corr/abs-2108-05080} and KoDF~\cite{Kwon_2021_ICCV} dataset across demographics.}
    \label{fig:img1-samples}
\end{figure}

\noindent \textbf{FakeAVCeleb:} The FakeAVCeleb dataset~\cite{DBLP:journals/corr/abs-2108-05080} is a collection of videos with audio and video manipulations of celebrities that have been generated using various deepfake techniques. The dataset is created by selecting videos from the VoxCeleb2~\cite{chung18b_interspeech} dataset, featuring $500$ celebrities. 
The dataset is well-balanced and annotated in terms of gender, race, geography, and visual and audio manipulations, making it useful for training deep learning models that can generalize well on unseen test sets.  
We chose this dataset for our experiments for multimodal detection because it contains both audio and visual manipulations, as well as a variety of deepfake generation techniques.  
We have used the gender and race-balanced version of the training and test set in this study. Figure~\ref{fig:img1-samples} shows sample frames from the datasets across demographics. \\

\noindent \textbf{KoDF:} The KoDF~\cite{Kwon_2021_ICCV} (Korean deepfake) dataset is a collection of synthetic and real videos of Korean celebrities that have been generated using various deepfake generation techniques. 
As this dataset has real and fake video and only real audio, it is used only for cross-dataset evaluation for the visual and multimodal deepfake detectors. We have used the gender-balanced version of the test set in this study. \\

\noindent \textbf{Implementation Details:}
\label{implementation}
The proposed model uses MTCNN~\cite{7553523} for face detection and Xception-based deep feature extraction from visual sequence and the Mel-frequency cepstrum coefficient (MFCC)~\cite{10.1145/3476099.3484315} based acoustic representation for audio sequence. 
MTCNN utilizes a cascaded framework for joint face detection and alignment. The images are then resized to $256\times256$ for both training and evaluation. The deep features were extracted from the fully connected layer of the pretrained Xception model.  We chose $300$ frames per video for training and $180$ frames per video for validation and testing of the models. 
MFCC is used for raw audio file conversion into feature vectors for the model training with a window size of $25$ms and a hop of $10$ms. The extracted frames from visual and MFCC are segmented into visual and audio sequences. 

The hyperparameters, including activation functions and dropout probabilities, were selected through a grid-search process using the validation set. The Adam optimizer with an exponential decay learning rate scheduler is used for optimization, and an early-stopping strategy is implemented with patience of $11$ epochs to determine the optimal training duration for each model. The final hyperparameters for the model are as follows: in modality-invariant loss, $CMD_K$ (refer Eq.~\ref{eq9}) is set to $5$, the activation function used is tanh for the LSTM layers and ReLU for all other layers, the gradient clip is $1.0$, the regularization parameters $\alpha$, $\beta$, and $\gamma$ (refer Eq.~\ref{eq8}) are set to $0.7$, $1.0$, and $0.7$ respectively, the dropout probability is $0.1$, the hidden dimension for the LSTM layers is $128$, the batch size is $32$, and the learning rate is set to $0.0001$. \\

\noindent \textbf{Evaluation Metrics}: For the performance evaluation, we employed standard evaluation metrics commonly used for deepfake detection, such as Area under the ROC Curve~(AUC), partial AUC (pAUC) (at $10$\% False Positive Rate~(FPR)), and Equal Error Rate~(EER). 
We followed the evaluation procedure established by the National Institute of Standards and Technology (NIST) and assessed the overall classification accuracy (ACC), along with the true positive rate~(TPR), and false-positive rate~(FPR) similar to the studies on the deepfake detectors~\cite{ Nadimpalli2022GBDFGB}.

\subsection{{\textbf{Baselines}}}
\label{baseline}
\begin{enumerate}
\item \textbf{Unimodal-Audio} and \textbf{-Visual:} These models use MFCC and deep  features extracted from audio and visual modalities, respectively, as described in Section~\ref{implementation} generating the feature representations \(x_{i}^{a}\) and \(x_{i}^{v}\) (refer Eq.~\ref{eq1}). These are given to the specific encoder to generate the hidden representations \(h_{i}^{a}\) and \(h_{i}^{v}\) (refer Eq.~\ref{eq2}). This is followed by four fully connected (dense) layers of size $1024$ and the final output layer with sigmoid activation for deepfake classification.
\item \textbf{Multimodal Deeplearning (feature concatenation):} The hidden feature representations (\(h_{i}^{a}\) and \(h_{i}^{v}\)) (refer Eq.~\ref{eq2}) from the specific encoders for both the visual and audio sequences are concatenated together followed by two dense layers of size $1024$, and the output sigmoid layer for deepfake detection. 

\item \textbf{Ensemble-Soft Voting:} The individual Unimodal-Audio and -Visual approaches for deepfake detection are combined into an ensemble using the soft voting-based average rule. The soft voting-based average rule uses a class-wise average of probability values obtained from each classifier in the ensemble for the final classification. 
 
\end{enumerate}

\subsection{\textbf{Results and Analysis}}
In this section, we examine the results of the deepfake detectors trained on the FakeAVCeleb and tested on the FakeAVCeleb and KoDF test sets.  All the \textbf{evaluation metrics} (from section~\ref{metrics}) are reported in the range $[0, 1]$. 

\begin{table}[ht]
\caption{Performance of the proposed MIS-AVoiDD when trained on FakeAVCeleb and tested on \textbf{FakeAVCeleb} and \textbf{KoDF}. Cross-comparison of the proposed method is done with unimodal audio and visual-based deepfake detectors and multimodal deepfake detectors based on feature concatenation and soft-voting schemes. The metrics used are AUC, pAUC, EER, ACC, TPR, and FPR.}
\label{table1}
\centering
\scalebox{0.8}{ 
\begin{tabular}{cl|c|c}
\hline
\multirow{2}{*}{\textbf{Model}} & \multicolumn{1}{c|}{\multirow{2}{*}{\textbf{Metric}}} & \multirow{2}{*}{\textbf{FakeAVCeleb}} & \multirow{2}{*}{\textbf{KoDF}} \\
 & \multicolumn{1}{c|}{} &  &  \\ \hline
\multirow{6}{*}{\textbf{MIS-AVoiDD}} & \textbf{AUC} & 0.973 & 0.962 \\
 & \textbf{pAUC} & 0.961 & 0.946 \\
 & \textbf{EER} & 0.039 & 0.044 \\
 & \textbf{ACC} & 0.962 & 0.950 \\
 & \textbf{TPR} & 0.943 & 0.935 \\
 & \textbf{FPR} & 0.047 & 0.059 \\ \hline
 \multirow{6}{*}{\textbf{\begin{tabular}[c]{@{}c@{}}Multimodal Deeplearning\\ (feature concatenation)\end{tabular}}} & \textbf{AUC} & 0.890 & 0.865 \\
 & \textbf{pAUC} & 0.861 & 0.848 \\
 & \textbf{EER} & 0.157 & 0.189 \\
 & \textbf{ACC} & 0.856 & 0.849 \\
 & \textbf{TPR} & 0.841 & 0.829 \\
 & \textbf{FPR} & 0.183 & 0.218 \\ \hline
 \multirow{6}{*}{\textbf{Ensemble-Soft Voting}} & \textbf{AUC} & 0.844
 &  0.817
\\
 & \textbf{pAUC} 
 & 0.838 & 0.787
 \\
 & \textbf{EER} & 0.192
 &  0.216
 \\
 & \textbf{ACC} & 0.832
 & 0.808
 \\
 & \textbf{TPR} & 0.821
 & 0.786
 \\
 & \textbf{FPR} & 0.232
 & 0.243
 \\ \hline
\multirow{6}{*}{\textbf{Unimodal-Audio}} & \textbf{AUC} & 0.813 & - \\
 & \textbf{pAUC} & 0.793 & - \\
 & \textbf{EER} & 0.237 & - \\
 & \textbf{ACC} & 0.801 & - \\
 & \textbf{TPR} & 0.784 & - \\
 & \textbf{FPR} & 0.257 & - \\ \hline
\multirow{6}{*}{\textbf{Unimodal-Visual}} & \textbf{AUC} & 0.809 & 0.783 \\
 & \textbf{pAUC} & 0.792 & 0.755 \\
 & \textbf{EER} & 0.250 & 0.274 \\
 & \textbf{ACC} & 0.800 & 0.778 \\
 & \textbf{TPR} & 0.785 & 0.758 \\
 & \textbf{FPR} & 0.261 & 0.282 \\ \hline
\end{tabular}}
\end{table}

Table~\ref{table1} shows the performance of the proposed approach based on modality invariant and specific features (MIS-AVoiDD) for deepfake detection. Cross-comparison is done with the baseline unimodal and multimodal audio and visual deepfake detectors (as discussed in section~\ref{baseline}) on FakeAVCeleb and KoDF test sets. 

For the FakeAVCeleb dataset, the MIS-AVoiDD model outperforms all the compared unimodal and multimodal deepfake detectors. The performance increment of the MIS-AVoiDD is by $0.083$, $0.100$, and $0.118$ in terms of AUC, pAUC, and EER over multimodal deep learning (feature concatenation) deepfake detector. MIS-AVoiDD has a performance increment of $0.129$, $0.123$, and $0.153$ in terms of  AUC, pAUC, and EER over the ensemble soft voting-based approach. Similar trends can be observed for other metrics,  consistently demonstrating the best performance of the proposed model. MIS-AVoiDD has a performance increment of $0.16$, $0.168$, and $0.198$ in terms of  AUC, pAUC, and EER over the Unimodal-Audio. Over Unimodal-Visual, our proposed approach has an increment of $0.164$, $0.169$, and $0.211$ in terms of  AUC, pAUC, and EER. Similar trends can be observed for other metrics,  consistently demonstrating the best performance of the proposed model.
The Unimodal-Audio and Unimodal-Visual models obtain similar performance with respect to the multimodal deep learning (feature concatenation) based model with the performance increment of $0.077$, $0.068$, and $0.080$ in terms of AUC, pAUC, and EER respectively.

Over the KoDF dataset, the MIS-AVoiDD model performs the best with about $0.097$, $0.098$, and $0.145$ performance improvement in terms of AUC, pAUC, and EER over the multimodal deep learning (feature concatenation) approach. MIS-AVoiDD has a performance increment of $0.146$, $0.159$, and $0.172$ in terms of  AUC, pAUC, and EER over the soft voting-based approach. The Unimodal-Audio is not applicable as the KoDF dataset does not have fake audio and is denoted by dashes (-) in the table.
MIS-AVoiDD has a performance increment of $0.179$, $0.191$, and $0.230$ in terms of  AUC, pAUC, and EER over the Unimodal-Visual.
The Unimodal-Visual model has a performance decrement of $0.077$, $0.068$, and $0.080$ in terms of AUC, pAUC, and EER respectively with respect to multimodal deep learning (feature concatenation).

Overall, MIS-AVoiDD consistently obtains superior performance compared to unimodal and multimodal detectors with an average performance increment of $0.135$, $0.175$, and $0.142$ in terms of AUC, EER, and ACC, respectively, on the FakeAVCeleb test set. For the KoDF test set the proposed model obtained an average performance increment of $0.138$, $0.187$, and $0.136$ in terms of AUC, EER, and ACC, respectively. 

\begin{table}[ht]
\caption{Overall comparison of MIS-AVoiDD with SOTA approaches tested on \textbf{FakeAVCeleb} dataset. The metrics used are AUC and ACC.}
\label{tablec}
\centering
\scalebox{0.91}{ 
\begin{tabular}{c|c|cc}
\hline
\multirow{2}{*}{\textbf{Model}} & \multirow{2}{*}{\textbf{Modality}} & \multicolumn{2}{c}{\textbf{FakeAVCeleb}} \\
 &  & \textbf{ACC (\%)} & \textbf{AUC} \\ \hline
 Head Pose~\cite{yang2019exposing} & Visual & 68.8 & 0.709 \\ 
 VA-MLP~\cite{matern2019exploiting} & Visual & 65 & 0.671 \\
 DeFakeHop~\cite{chen2021defakehop} & Visual & 68.3 & 0.716 \\
 CViT ~\cite{wodajo2021deepfake} & Visual & 69.7 & 0.718 \\
 Multiple attention ~\cite{zhao2021multi} & Visual & 77.6 & 0.793 \\
 SLADD~\cite{chen2022self} & Visual & 70.5 & 0.721 \\
 Lip Forensics~\cite{9980296} & Visual & 80.1 & 0.824 \\ \hline
 TDNN~\cite{pianese2022deepfake} & Audio & 59.8 & 0.627 \\
 LFCC and LCNN ~\cite{monteiro2019end} & Audio & 47.4 & 0.503 \\
 MFCC + VGG16~\cite{hamza2022deepfake} & Audio & 67.14 & 0.671 \\
 MFCC + Xception~\cite{khalid2021fakeavceleb} & Audio & 76.26 & 0.762 \\
 Lip Forensics~\cite{9980296} & Audio & 60.0 & - \\
 \hline
AVN-J~\cite{9350195} & Audio-Visual & 73.2 & 0.776 \\
MDS~\cite{10.1145/3394171.3413700} & Audio-Visual & 82.8 & 0.865 \\
Emotions don't lie ~\cite{10.1145/3394171.3413570}  & Audio-Visual & 78.1 & 0.798 \\
AVFAkeNet~\cite{ILYAS2023110124} & Audio-Visual  & 78.4 & 0.834 \\
AVoiD~\cite{yang2023avoid} & Audio-Visual & 83.7 & 0.892 \\ 
Multimodal-1 ~\cite{10.1145/3476099.3484315} & Audio-Visual & 50.0 & - \\
Multimodal-2 ~\cite{10.1145/3476099.3484315} & Audio-Visual & 67.4 & - \\
AVTS-DFD ~\cite{10095247} & Audio-Visual & 94.4 & - \\ 
Lip-Sync~\cite{9980296} & Audio-Visual & 83.3 & 0.976 \\
AV-POI~\cite{Cozzolino2022AudioVisualPD} & Audio-Visual & 86.6 & 0.942 \\
VFD~\cite{cheng2022voice} & Audio-Visual & 81.5 & 0.861 \\
AVForensics~\cite{feng2023self} & Audio-Visual & - & 0.900 \\
MultimodalTrace~\cite{raza2023multimodaltrace} & Audio-Visual & 92.9 & - \\
\hline
\textbf{MIS-AVoiDD (Ours)} & Audio-Visual & \textbf{96.2} & \textbf{0.973} \\ \hline
\end{tabular}}
\end{table}

Table~\ref{tablec} compares MIS-AVoiDD with SOTA unimodal and multimodal deepfake detectors when tested on the FakeAVCeleb dataset. In comparison to the visual modality-based deepfake detectors, on average MIS-AVoiDD obtained a performance improvement of $16.1$\% and $0.151$ in terms of ACC and AUC respectively. In comparison to the audio-based deepfake detectors, on average MIS-AVoiDD has obtained performance improvement of $20.06$\% and $0.200$ in terms of ACC and AUC. 
 

With respect to all the multimodal audio-visual deepfake detectors (Table~\ref{tablec}), our proposed MIS-AVoiDD  significantly outperforms recently published  AVoiD~\cite{yang2023avoid}, AVFakeNet ~\cite{ILYAS2023110124}, AVTS-DFD ~\cite{10095247}, Lip-Sync~\cite{9980296}, AV-POI~\cite{Cozzolino2022AudioVisualPD}, VFD~\cite{cheng2022voice}, and multimodal approach~\cite{10.1145/3476099.3484315} with an average performance increment of $0.184$ and $0.102$ in terms of ACC and AUC. Overall, compared to existing multimodal methods, MIS-AVoiDD obtains state-of-the-art performance.  

Worth discussing, the audio-visual deepfake detector proposed in~\cite{raza2023multimodaltrace} also handles the problem at the representation level by projecting audio-visual streams into shared and individual spaces represented by MLP blocks. However, the MIS-AVoiDD approach surpassed~\cite{raza2023multimodaltrace} with an accuracy increment of $3.4$\%. This could be attributed to the use of an attention mechanism for feature fusion. Very recently, a study in~\cite{chen2023npvforensics} has proposed a multimodal deepfake detector. However, this detector uses three modalities i.e., face, audio, and the extracted lip sequences. Therefore, this study is not used for the comparison with our proposed model.

In \textbf{summary}, our experimental results suggest the better performance of MIS-AVoiDD when compared to the unimodal and the multimodal baseline approaches by about $0.134$, $0.14$, and $0.17$ in terms of AUC, pAUC, and EER. 


\section{Ablation Study}
\label{ablation}
An ablation study is conducted on MIS-AVoiDD by varying the regularization parameters of the loss functions (see Eq.~\ref{eq8}) and feature representation subspaces i.e., modality invariant and modality-specific subspaces (refer Section~\ref{representation}). 
\subsection{\textbf{Significance of Regularization}}
Regularization plays a critical role in achieving the desired representations discussed in Section~\ref{regularization} and Eq.~\ref{eq8}. We looked at the importance of each loss by an ablation study. To quantitatively verify the importance of these losses, we re-trained the model by ablating one loss at a time. To nullify each
loss, we set either {$\alpha$, $\beta$, $\gamma$} to $0$ in Eq.~\ref{eq8}. Results are observed in Table~\ref{table4}. 

As seen, the best performance is achieved when all the losses are involved. In a closer look, we can see that the models are particularly sensitive to the orthogonal ($\mathcal{L}_{\text{Orth}}$ in Eq.~\ref{eq10}) and invariant losses ($\mathcal{L}_{\text{Inv}}$ in Eq.~\ref{eq9}) that ensure both the modality invariance and specificity. MIS-AVoiDD with all the regularization parameters when compared to ablating orthogonal loss ($\mathcal{L}_{\text{Orth}}$ in Eq.~\ref{eq10}), has a performance loss of $0.047$, $0.055$, $0.081$, $0.05$, $0.054$, and $0.099$ in terms of AUC, pAUC, EER, ACC, TPR, and FPR. This dependence indicates that having separate subspaces is indeed helpful. MIS-AVoiDD compared to ablating invariant loss has a performance increment of $0.031$, $0.036$, $0.074$, $0.03$, $0.025$, and $0.08$ in terms of AUC, pAUC, EER, ACC, TPR, and FPR.

For the similarity loss ($\mathcal{L}_{\text{Sim}}$ in Eq.~\ref{eq11}), we see a lesser dependence on the model.
There is a performance decrement of $0.013$, $0.018$, $0.044$, $0.016$, $0.01$, and $0.057$ in terms of AUC, pAUC, EER, ACC, TPR, and FPR. One possibility is that, despite the absence of similarity loss, the modality-specific encoders are not resorting to trivial solutions but rather learning informative representations using the classification loss. This would not be the case if only the modality-invariant features were used for prediction. The ablation study suggests the inclusion of all the regularization roles leads to improved performance in deepfake detection.

\begin{table}[ht]
\caption{Evaluation of MIS-AVoiDD for different roles of regularization (ablating invariant, orthogonal, and similarity losses) and subspaces (specific and invariant). (-) in the table represents the removal of the regularization. The metrics used are AUC, pAUC, EER, ACC, TPR, and FPR.}
\label{table4}
\centering
\scalebox{0.77}{
\begin{tabular}{cc|l|l|l|l|l|l}
\hline
\multicolumn{2}{c|}{\textbf{Modal}}                           & \textbf{AUC} & \textbf{pAUC} & \textbf{EER} & \textbf{ACC} & \textbf{TPR} & \textbf{FPR} \\ \hline
\multicolumn{1}{c|}{\multirow{3}{*}{\textbf{Regularization}}} & \textbf{\textbf{(-)$\mathcal{L}_{\text{Inv}} (\alpha = $0$) $}} & 0.942 & 0.925 & 0.113 & 0.932 & 0.918 & 0.127 \\  
                                        \multicolumn{1}{c|}{} & \textbf{\textbf{(-)$\mathcal{L}_{\text{Orth}} (\beta=$0$)$}} & 0.926 & 0.906 & 0.124 & 0.912 & 0.889 & 0.146 \\  
                                        \multicolumn{1}{c|}{} & \textbf{\textbf{(-)$\mathcal{L}_{\text{Sim}}(\gamma=$0$)$}} & 0.960 & 0.943 & 0.083 & 0.946 & 0.933 & 0.104 \\ \hline
\multicolumn{1}{c|}{\multirow{2}{*}{\textbf{Subspaces}}}      & \textbf{Modality Specific}  & 0.949 & 0.931 & 0.106 & 0.932 & 0.917 & 0.118 \\ 
                                       \multicolumn{1}{c|}{}  & \textbf{Modality Invariant} & 0.964 & 0.950 & 0.072 & 0.947 & 0.933 & 0.085              \\ \hline
\multicolumn{2}{c|}{\textbf{MIS-AVoiDD}}                                                     & 0.973 & 0.961 & 0.039 & 0.962 & 0.943 & 0.047 \\ \hline
\end{tabular}}
\end{table}

\subsection{\textbf{Significance of Subspaces}}
In this section, we look at two variants of our proposed model to investigate the significance of subspaces (i.e., modality-specific and invariant subspaces, refer to Section~\ref{representation}). In the modality-specific variation, only specific feature representations are used for fusion and further prediction. While in the modality invariant variation, only invariant feature representations are used for fusion and further prediction. 

From the results, it is evident that the model has performed best with both the specific and invariant features. Having only the specific subspace is more sensitive when compared to having only the invariant subspace. 
Specific subspace has a performance decrement of $0.024$, $0.03$, $0.067$, $0.03$, $0.026$, and $0.071$ in terms of AUC, pAUC, EER, ACC, TPR, and FPR. The invariant subspace has a performance decrement of $0.009$, $0.011$, $0.033$, $0.015$, $0.01$, and $0.038$ in terms of AUC, pAUC, EER, ACC, TPR, and FPR. The superior performance of invariant subspace suggests the importance of common patterns and inconsistencies between the audio and visual streams in multimodal deepfake detection. 
\section{Conclusion}
\label{conclusion}
With the staggering growth in deepfake generation techniques, multimodal deepfakes, in the form of forged videos with lip-synced synthetic audios, have emerged as a novel threat. Consequently, a new generation of multimodal audio-visual deepfake detectors is being investigated to detect audio and visual manipulations collectively. Current multimodal deepfake detectors are often based on the fusion of audio and visual streams. However, due to the heterogeneous nature of the audio and visual streams, there is a need for an advanced mechanism for multimodal manipulation detection. In this paper, we proposed the use of multimodal representation learning capturing both modality-specific and invariant patterns for joint audio-visual deepfake detection. Our experimental results suggest the enhanced performance of our proposed model over existing unimodal and multimodal deepfake detectors, obtaining SOTA performance. Thus, suggesting the importance of common patterns and patterns specific to each modality representing pristine or fake content for audio-visual manipulation detection collectively.


\small{
\balance
\bibliographystyle{IEEE}
\bibliography{refs}}
\end{document}